\documentclass[sigconf]{acmart}
\usepackage[linesnumbered,boxed,ruled,commentsnumbered]{algorithm2e}
\usepackage{graphicx}
\usepackage{amsmath}
\usepackage{booktabs} 
\usepackage{color}
\definecolor{ao}{rgb}{0.0, 0.5, 0.0}
\usepackage{comment}
\usepackage{booktabs}
\usepackage{amsfonts}
\usepackage{amsmath}

\usepackage{amssymb}
\usepackage{makecell}
\usepackage{subcaption}
\usepackage{multirow}
\usepackage{pbox}
\usepackage{hhline}
\usepackage{graphicx}
\usepackage[flushleft]{threeparttable}
\usepackage{svg}
\usepackage{courier}
\usepackage{xspace}
\usepackage{epstopdf}
\usepackage{stfloats}
\usepackage{array}
\usepackage{xcolor,colortbl}
\colorlet{shadecolor}{gray!20}

\usepackage{mleftright}

\usepackage[colorinlistoftodos]{todonotes}

\usepackage{bm}
\usepackage{bbm}

\usepackage[aboveskip=1pt]{subcaption}
\usepackage[skip=0.85\baselineskip]{caption}
\usepackage{makecell}

\AtBeginDocument{%
  \providecommand\BibTeX{{%
    \normalfont B\kern-0.5em{\scshape i\kern-0.25em b}\kern-0.8em\TeX}}}

\copyrightyear{2020}
\acmYear{2020}
\setcopyright{rightsretained}

\acmConference[KDD-DLG '20]{the 26th International Conference on Knowledge Discovery and Data Mining}{August 2020}{San Diego, California, USA}
\begin{document}

\title{Hierarchical Attention Models for Multi-Relational Graphs}


\author{Roshni G. Iyer}
\affiliation{%
  \institution{University of California, Los Angeles}
  \city{Los Angeles}
  \state{CA}
  \country{USA}
}
\email{roshnigiyer@cs.ucla.edu}

\author{Wei Wang}
\affiliation{%
  \institution{University of California, Los Angeles}
  \city{Los Angeles}
  \state{CA}
  \country{USA}
 }
\email{weiwang@cs.ucla.edu}

\author{Yizhou Sun}
\affiliation{%
  \institution{University of California, Los Angeles}
  \city{Los Angeles}
  \state{CA}
  \country{USA}
}
\email{yzsun@cs.ucla.edu}





\renewcommand{\shortauthors}{Roshni G. Iyer et al.}
\newcommand{\ourmodel}{\textbf{\textsc{DGS}}\xspace}
\newcommand{\transe}{\textbf{\textsc{TransE}}\xspace}
\newcommand{\distmult}{\textbf{\textsc{DistMult}}\xspace}
\newcommand{\complex}{\textbf{\textsc{ComplEx}}\xspace}
\newcommand{\rotate}{\textbf{\textsc{RotatE}}\xspace}
\newcommand{\joie}{\textbf{\textsc{JOIE}}\xspace}
\newcommand{\hyperkg}{\textbf{\textsc{HyperKG}}\xspace}
\newcommand{\hake}{\textbf{\textsc{HAKE}}\xspace}
\newcommand{\cone}{\textbf{\textsc{ConE}}\xspace}
\newcommand{\rehf}{\textbf{\textsc{RefH}}\xspace}
\newcommand{\roth}{\textbf{\textsc{RotH}}\xspace}
\newcommand{\atth}{\textbf{\textsc{AttH}}\xspace}
\newcommand{\hgcn}{\textbf{\textsc{HGCN}}\xspace}
\newcommand{\hyperka}{\textbf{\textsc{HyperKA}}\xspace}
\newcommand{\mtwognn}{\textbf{\textsc{\textrm{M$^{2}$GNN}}}\xspace}
\newcommand{\murp}{\textbf{\textsc{MuRP}}\xspace}

\newcommand\norm[1]{\lVert#1\rVert}

\begin{abstract}
We present Bi-Level Attention-Based Relational Graph Convolutional Networks (BR-GCN), unique neural network architectures that utilize masked self-attentional layers with relational graph convolutions, to effectively operate on highly multi-relational data. BR-GCN models use bi-level attention to learn node embeddings through (1) node-level attention, and (2) relation-level attention. The node-level self-attentional layers use intra-relational graph interactions to learn relation-specific node embeddings using a weighted aggregation of neighborhood features in a sparse subgraph region. The relation-level self-attentional layers use inter-relational graph interactions to learn the final node embeddings using a weighted aggregation of relation-specific node embeddings. The BR-GCN bi-level attention mechanism extends Transformer-based multiplicative attention from the natural language processing (NLP) domain, and Graph Attention Networks (GAT)-based attention, to large-scale heterogeneous graphs (HGs).  On node classification, BR-GCN outperforms baselines from 0.29\% to 14.95\% as a stand-alone model, and on link prediction, BR-GCN outperforms baselines from 0.02\% to 7.40\% as an auto-encoder model. We also conduct ablation studies to evaluate the quality of BR-GCN's relation-level attention and discuss how its learning of graph structure may be transferred to enrich other graph neural networks (GNNs). Through various experiments, we show that BR-GCN's attention mechanism is both scalable and more effective in learning compared to state-of-the-art GNNs.
\end{abstract}

\keywords{information retrieval, graph neural networks, graph attention}


\maketitle

\section{BR-GCN Architecture}
\label{BR-GCN-architecture}

\begin{table*}[t!]
  \caption{\textmd{Variables (Var) and Explanations. Table on the left column corresponds to node-level attention. Table on the right column corresponds to relation-level attention.}}
  \label{tab:dataset_description}
  \centering
  \begin{tabular}{cc|cc}
  \hline
Var & Explanation & Var & Explanation \\ [0.5 ex]
\hline
$r$ & Relation on node edge & $\boldsymbol{W}_{1, r}$ & Projection weight matrix for $\boldsymbol{z}_{i}^{r}$\\[0.5ex]
$R_{i}$ & Set of relations on edge of node $i$ & $\boldsymbol{W}_{2, r}$ & Projection weight matrix for $\boldsymbol{z}_{i}^{r}$\\[0.5ex] 
$\boldsymbol{h}_{i}^{(l)}$ & Node $i$ features at layer $l$ & $\boldsymbol{W}_{3, r}$ & Projection weight matrix for $\boldsymbol{z}_{i}^{r}$\\[0.5ex]
$e_{i,j}^{r}$ & GAT-based attention for $(i,j)$ & $\boldsymbol{q}_{r, i} \in Q_{r}$  & Transformer-based query matrix row\\[0.5ex]
$\mathrm{edge}_{i, j}$ & Edge between node $i$ and node $j$ & $\boldsymbol{k}_{r, i} \in K_{r}$ & Transformer-based key matrix row\\[0.5ex]
$\boldsymbol{a}_{r}$ & Relation-specific attention vector & $\boldsymbol{v}_{r, i} \in V_{r}$ & Transformer-based value matrix row\\[0.5ex]
$\gamma_{i,j}^{r}$ & Relation-specific weight for $(i,j)$ & $\boldsymbol{W}_{i}$ & Weight matrix for $\boldsymbol{h}_{i}^{(l)}$\\[0.5ex]
$N_{i}^{r}$ &  Set of relation-specific node neighbors & $\psi_{i}^{r, r^{\prime}}$ & Learned attention for relation pairs ($r$, $r^{\prime}$)\\[0.5ex]
$\boldsymbol{z}_{i}^{r}$ &  Relation-specific node embedding & $\boldsymbol{\delta}_{i}^{r}$ & Attended relation-specific embedding\\[0.5ex]
\hline
 \end{tabular}
\end{table*}

We define directed and labeled HGs as utilized in this work as $\mathcal{G} = (\mathcal{V}, \mathcal{E}, \mathcal{R})$ where nodes are $v_{i} \in \mathcal{V}$ and belong to possibly different entities, and edges are $(v_{i}, r, v_{j}) \in \mathcal{E}$ with $r \in \mathcal{R}$ and belong to possibly different relation types. 

\subsection{Generalized Framework for Computing Bi-Level Attention} 

Bi-level attention is more powerful in learning compared to uni-level attention, where only one level of attention is learned by the model. Bi-level attention learns attention at different levels of granularity in the HG which thereby captures more information about graph components than a uni-level attention mechanism is capable of. Eq.~\ref{eqn:BR-GCN-overview-attention} describes the generalized bi-level attention framework to compute embeddings for node $i$ in the $(l+1)$-th layer: 
\vspace{-3mm}
\begin{multline}
\label{eqn:BR-GCN-overview-attention}
\boldsymbol{h}_{i}^{(l+1)} = \mathrm{AGG}(\Big\{\boldsymbol{f}(a(r))^{T}\mathrm{AGG}(\Big\{\\\boldsymbol{g}(a(\mathrm{edge}_{i,j}|r))|j \in N_{i}^{r}\Big\})|r \in R_{i}\Big\})
\end{multline}
where $\boldsymbol{g}(a(\mathrm{edge}_{i,j}|r))$ is a vector-output function of the node-level attention that provides a relation-specific embedding summary which is aggregated, $\mathrm{AGG}(\cdot)$, over edges $\mathrm{edge}_{i,j}$ that belong to the neighborhood context of nodes $j \in N_{i}^{r}$. $\boldsymbol{f}(a(r))$ is a vector-output function of the relation-level attention that are weighted relation-specific embeddings which are aggregated over relations in the neighborhood context to form the final node embedding. See Table~\ref{tab:dataset_description} for explanations of \vspace{2mm} variables. 

\begin{figure*}[htp]
    \centering
    \includegraphics[width=12cm]{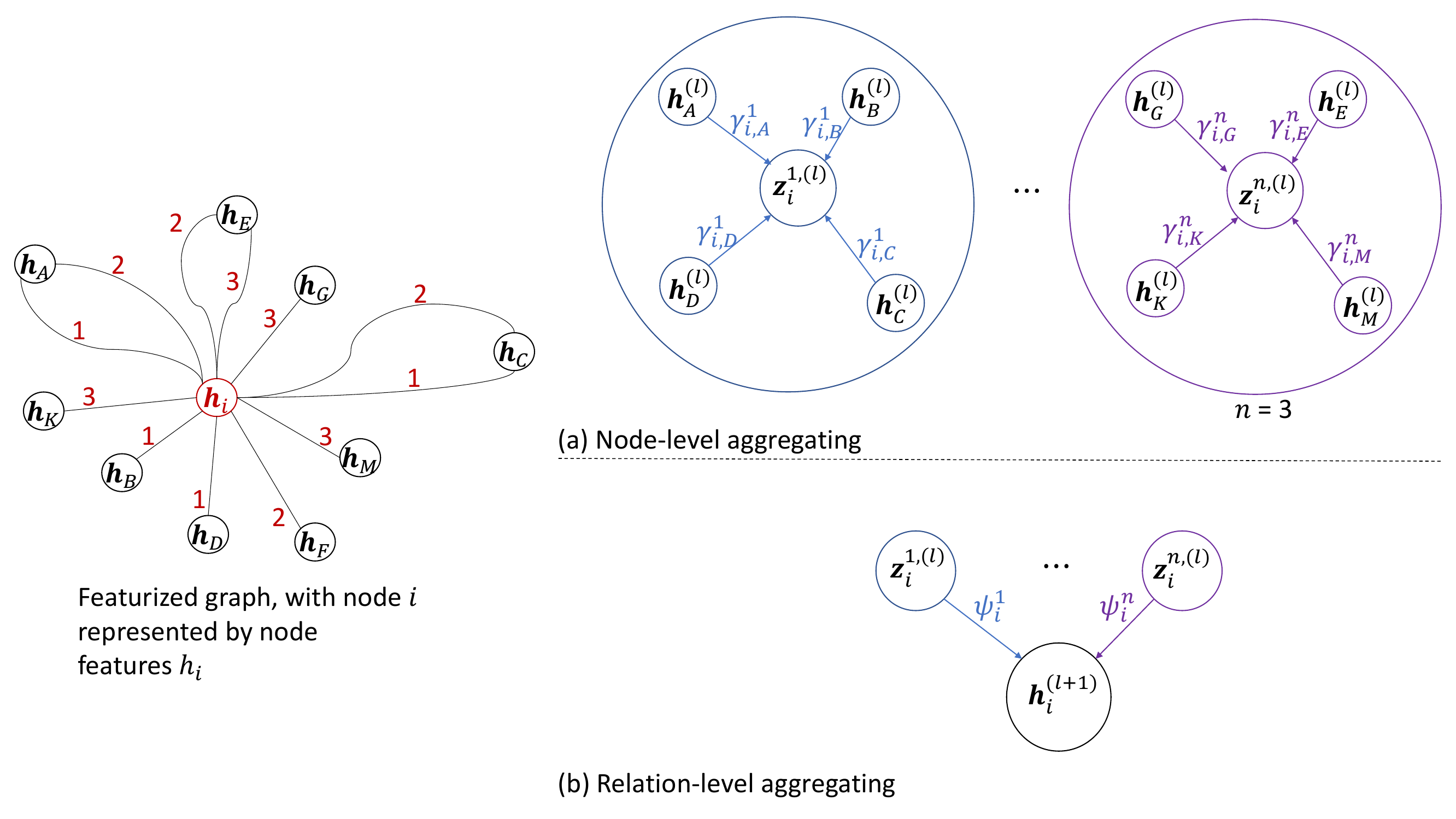}
    \caption{\textmd{Bi-level attention visualization. (a) Node-level aggregating: A node’s features in the ($l+1$)-th layer is a weighted combination of relation-specific embeddings of the node, $\boldsymbol{z}_{i}^{r}$. (b) Relation-level aggregating:  Relation-level attention is learned through multiplicative attention using neighborhood relational similarity to other relations to determine the relation's relative importance.}}
    \label{fig:2l-attention}
\end{figure*}

In Sections~\ref{subsection:node-level-att} and~\ref{subsection:relation-level-att}, we propose a novel semi-supervised attention-based GCN model, BR-GCN, for multi-relational HGs. BR-GCN models use bi-level attention to learn (1) node-level attention, followed by (2) relation-level attention. BR-GCN's attention mechanism is summarized in Figure~\ref{fig:2l-attention}. BR-GCN models use $L$ stacked layers, each of which is defined through Eq.~\ref{eqn:BR-GCN-overview-attention}, where the previous layer's output is input to the next layer. The initialized input can be chosen as a unique one-hot vector for each node if no other features are present. The model also supports pre-defined features. BR-GCN's source code, pseudo-code, and a walkthrough example of its attention mechanism is in the Appendix. 

\subsection{Node-level Attention}
\label{subsection:node-level-att}
Node-level attention distinguishes the different roles of nodes in the neighborhood context for learning relation-specific node embeddings. As node-level attentions are target-node-specific, they are different for different target nodes. In HGs, neighbor nodes may belong to different feature spaces, so the features of all nodes are projected to the same feature space to enable node-level attention to handle arbitrary node types. BR-GCN's node-level attention uses additive attention similar to GAT~\cite{GAT}, but overcomes GAT's limitation by extending the attention to HGs. The self-attention for pair $(i, j)$ for node $j$'s importance to node $i$ for relation $r$ is defined as:
\begin{equation}
    \label{eqn:BR-GCN-e_ij}
    e_{i,j}^{r} = a(\mathrm{edge}_{i,j}|r) = att_{node}(\boldsymbol{h}_{i}^{(l)}, \boldsymbol{h}_{j}^{(l)}, r) 
\end{equation}
\begin{equation}
    \label{eqn:BR-GCN-with_LeakyReLU}
    = \mathrm{LeakyReLU}(\boldsymbol{a}_{r}^{T^{(l)}}\left[\boldsymbol{h}_{i}^{(l)}\parallel \boldsymbol{h}_{j}^{(l)}\right])
\end{equation}
For a specific relation $r$, $att_{node}(\cdot)$ is shared for all
node pairs, so that each node is influenced by its neighborhood context. $e_{i,j}^{r}$ is asymmetric since the importance of node $j$ to node $i$ may be different from the importance of node $i$ to node $j$. $\boldsymbol{a}_{r}^{T}$ attends over the concatenated, $||$, node features of nodes $i$ and $j$ with an applied $\mathrm{LeakyReLU}(\cdot)$ activation. \vspace{1mm}


By restricting the attention to within the relation-specific neighborhood context of nodes $j \in N_{i}^{r}$, sparsity structural information is injected into the model through masked self-attentional layers. A $\mathrm{softmax}(\cdot)$ activation is then applied to normalize each node-pair attention weight:
\begin{equation}
\label{eqn:BR-GCN-node-level-attention}
\gamma_{i,j}^{r} = \mathrm{softmax}(e_{i,j}^{r})
\end{equation}
\begin{equation}
\label{eqn:BR-GCN-node-level-attention-details}
= \frac{\mathrm{exp}(\mathrm{LeakyReLU}(\boldsymbol{a}_{r}^{T^{(l)}}\left[\boldsymbol{h}_{i}^{(l)}\parallel \boldsymbol{h}_{j}^{(l)}\right]))}{\sum_{k \in N_{i}^{r}}\mathrm{exp}(\mathrm{LeakyReLU}(\boldsymbol{a}_{r}^{T^{(l)}}\left[\boldsymbol{h}_{i}^{(l)}\parallel \boldsymbol{h}_{k}^{(l)}\right]))}
\end{equation}
Node $i$'s relation-specific embedding, $\boldsymbol{z}_{i}^{r}$, can then be learned with $\mathrm{AGG}(\cdot)$ in Eq.~\ref{eqn:BR-GCN-overview-attention} being a weighted summation of the neighbor’s projected features as follows:
\begin{equation}
    \label{eqn:BR-GCN-semantic-specific-embedding}
    \boldsymbol{z}_{i}^{r} = \sum_{j \in N_{i}^{r}}[\boldsymbol{g}(a(\mathrm{edge}_{i,j}|r))] = \sum_{j \in N_{i}^{r}}[\gamma_{i,j}^{r}\boldsymbol{h}_{j}^{(l)}]
\end{equation}
where $\boldsymbol{z}_{i}^{r}$ serves as a summary of relation $r$ for node $i$.

\subsection{Relation-level Attention}
\label{subsection:relation-level-att}
Relation-level attention distinguishes the different roles of relations in the neighborhood context for learning more comprehensive node embeddings. In HGs, different relations may play different roles of importance for a node $i$, in addition to its relation-specific neighbor nodes. As such, we learn relation-level attention to better fuse node $i$'s relation-specific node embeddings. We extend Transformer-based multiplicative attention to the HG domain to learn relation-level attention by capturing the importance of a relation $r$ based on how similar it is to the other relations in the local neighborhood context. By restricting the set of relations to the local neighborhood, $r \in R_{i}$, we utilize multiplicative attention. Node $i$'s relation-specific Transformer-based query vector $\boldsymbol{q}_{r, i}$, key vector $\boldsymbol{k}_{r, i}$, and value vector $\boldsymbol{v}_{r, i}$ are computed as follows:   
\begin{equation}
\label{eqn:BR-GCN-Transformer-QKV}
\boldsymbol{q}_{r, i}; \boldsymbol{k}_{r, i}; \boldsymbol{v}_{r, i} = \boldsymbol{W}_{1,r}\boldsymbol{z}_{i}^{r}; \boldsymbol{W}_{2,r}\boldsymbol{z}_{i}^{r}; \boldsymbol{W}_{3,r}\boldsymbol{z}_{i}^{r}
\end{equation}
where $\boldsymbol{z}_{i}^{r}$ is projected onto the learnable weight matrices of \\ $\boldsymbol{W}_{1,r}, \boldsymbol{W}_{2,r}, \boldsymbol{W}_{3,r}$.

\vspace{1mm}
The relation-level attention for relation pairs $(r, r^{\prime})$ are computed by iterating over relations in the neighborhood context, $r^{\prime} \in R_{i}$. The importance of relation $r^{\prime}$ of node $i$ is denoted as follows with Eq.~\ref{eqn:BR-GCN-semantic-level} capturing relation similarity:
\begin{equation}
    \label{eqn:BR-GCN-semantic-level}
    \psi_{i}^{r, r^{\prime}} = a(r) = att_{relation}(\boldsymbol{z}_{i}^{r}, r) =\mathrm{softmax}(\boldsymbol{q}_{r, i}^{T}\boldsymbol{k}_{r^{\prime}, i})
\end{equation}
where the more similar $r^{\prime}$ is to $r$, the greater the attention weights of $r^{\prime}$, which results in more contribution of $r^{\prime}$'s embedding to node $i$'s final embedding. As in node-level attention, a $\mathrm{softmax}(\cdot)$ activation is applied to normalize each relation-pair attention weight. \vspace{1mm }

Similar to Realtional Graph Convolutional Networks (R-GCN)~\cite{rgcn}, to enable the representation of a node to be informed by its representation in previous layers, we add a self-connection of a special relation type to each node, which is projected onto $\boldsymbol{W}_{i}$, and aggregated to the attended relation-specific embedding, $\psi_{i}^{r, r^{\prime}}\boldsymbol{v}_{r^{\prime},i}$. Consistent with R-GCN's architecture, a $\mathrm{ReLU}(\cdot)$ activation is then applied to this aggregated embedding:
\begin{equation}
    \boldsymbol{\delta}_{i}^{r} = \mathrm{ReLU}(\sum_{r^{\prime} \in R_{i}}\psi_{i}^{r, r^{\prime}}\boldsymbol{v}_{r^{\prime},i} + \boldsymbol{W}_{i}\boldsymbol{h}_{i}^{(l)})
\end{equation}
The final node embedding for node $i$ is learned with $\mathrm{AGG}(\cdot)$ in Eq.~\ref{eqn:BR-GCN-overview-attention} being a weighted summation of the relation-specific embeddings:
\begin{equation}
    \label{eqn:BR-GCN-final-embedding}
    \boldsymbol{h}_{i}^{(l+1)} = \sum_{r \in R_{i}}[\boldsymbol{f}(a(r))] = \sum_{r \in R_{i}} [\boldsymbol{\delta}_{i}^{r}]
\end{equation}
Finally, putting the above equation components together, the final node embedding is learned by:
\begin{equation}
    \boldsymbol{h}_{i}^{(l+1)} = \sum_{r \in R_{i}} \mathrm{ReLU}(\sum_{r^{\prime} \in R_{i}}\mathrm{softmax}(\boldsymbol{q}_{r, i}^{T}\boldsymbol{k}_{r^{\prime}, i})\boldsymbol{v}_{r^{\prime}, i} + \boldsymbol{W}_{i}\boldsymbol{h}_{i}^{(l)})
\end{equation}

\subsection{Justification for BR-GCN's Bi-Level Attention Mechanism}
Our approach uses GAT-based attention to learn node-level attention because it is an effective additive attention mechanism, and using multiplicative attention requires the strong assumption that the importance of a node is a function of its similarity to other nodes in its context. There may be relation-specific nodes that are highly similar to each other but require different node-level attentions to be learned. Transformer-based attention is used to learn relation-level attention because it is an effective multiplicative attention mechanism, and since concatenation is not enough to capture relational importance. Relation features are characterized simply by their relation types, whereas node features may have several attributes. As such, the assumption of multiplicative attention is applicable to the relation-level unlike the node-level where it is more difficult to learn the similarity of nodes when many attribute factors are at play. By hierarchically learning node-level and relation-level attention of HGs, BR-GCN models address the limitations of R-GCN and GAT, since bi-level attention captures more information than uni-level attention. By considering the entire HG instead of subgraphs from pre-selected meta-paths as in Heterogeneous Graph Attention Networks (HAN)~\cite{HAN}, BR-GCN comprehensively learns different aspects of nodes through the entire set of nodes and relations. Furthermore, by learning relation-level attention using information from the neighborhood context instead of a generic global vector as in HAN, BR-GCN learns a more personalized attention for that relation to construct the final node embedding. 
\vspace{0.3mm}

\vspace{-3mm}
\section{Experiments}
\label{evaluation}
 In this section, we present our experiments on node classification and link prediction using the benchmark datasets: AIFB~\cite{RDF}, MUTAG~\cite{RDF}, BGS~\cite{RDF}, AM~\cite{RDF}, FB15k~\cite{FB15k}, WN18~\cite{WIN18RR}, and FB15k-237~\cite{FB15k-237}. The models evaluated are: BR-GCN and variant models, HAN, R-GCN and variant models, GAT, and Graph Isomorphism Networks (GIN)~\cite{GIN}. We also discuss our experiments on ablation studies to evaluate the quality of BR-GCN's relation-level attention and explain how its learning of graph structure may be transferred to enrich other GNNs. 
 \begin{table*}[t!]
  \caption{\textmd{Node classification test accuracy \%. Results are averaged over 10 runs and with benchmark splits from~\cite{rgcn}. Baseline models are  
  HAN~\cite{HAN}, R-GCN~\cite{rgcn}, GAT~\cite{GAT}, and GIN~\cite{GIN}. BR-GCN-node and BR-GCN-relation are uni-level attention models such that BR-GCN-node uses BR-GCN's node-level attention, and BR-GCN-relation uses BR-GCN's relation-level attention. BR-GCN's model is described in Section~\ref{BR-GCN-architecture}. Experiments are run using the PyTorch Geometric framework~\cite{pytorch-geometric} on an NVIDIA Tesla V100 GPU cluster.}}
  \label{table:entity-classificaton-results}
  \centering
  \begin{tabular}{ccccc}
  \hline
 Model & AIFB & MUTAG & BGS & AM \\ [0.5 ex]
\hline
 HAN & 96.68 $\pm$ 0.04 & 78.46 $\pm$ 0.07 & 86.84 $\pm$ 0.21 & 90.68 $\pm$ 0.23\\[0.5ex] 
 R-GCN & 95.83 $\pm$ 0.62 & 73.23 $\pm$ 0.48 & 83.10 $\pm$ 0.80 & 89.29 $\pm$ 0.35\\[0.5ex] 
 GAT & 92.50 $\pm$ 0.29 & 66.18 $\pm$ 0.00 & 77.93 $\pm$ 0.17 & 88.52 $\pm$ 1.65\\[0.5ex] 
 GIN & 96.18 $\pm$ 0.53 & 78.89 $\pm$ 0.09 & 86.42 $\pm$ 0.35 & 91.33 $\pm$ 0.16\\[0.5ex] 
 BR-GCN-node & 96.46 $\pm$ 0.13 & 73.19 $\pm$ 0.25 & 84.23 $\pm$ 0.22 & 89.45 $\pm$ 0.02\\[0.5ex] 
 BR-GCN-relation & 95.28 $\pm$ 0.23 & 76.17 $\pm$ 0.22 & 87.53 $\pm$ 0.34 & 90.52 $\pm$ 0.18\\[0.5ex] 
 BR-GCN (ours) & \textbf{96.97} $\pm$ 0.08 & \textbf{81.13} $\pm$ 0.61 & \textbf{88.30} $\pm$ 0.04 & \textbf{92.57} $\pm$ 0.15\\ [0.5ex] 
 \hline
 \end{tabular}
\end{table*}

\subsection{Node Classification}
Node classification is the semi-supervised classification of nodes to entity types. For evaluation consistency against R-GCN and GAT, BR-GCN architectures are implemented using two convolutional layers with the output of the final layer using a $\mathrm{softmax}(\cdot)$ activation per node. We optimize BR-GCN using cross-entropy loss for labeled nodes with parameters learned through the Adam Optimizer. We minimize the cross-entropy loss below:
\begin{equation}
    L = -\sum_{i \in Y}\sum_{k = 1}^{K}t_{ik}\mathrm{ln}h_{ik}^{(L)}
\end{equation}
with $Y$ being the indices of labeled nodes, $h_{ik}^{(L)}$ being the $k$-th entry of the $i$-th labeled node for the $L$-th layer, and $t_{ik}$ being $h_{ik}$'s corresponding ground-truth label.

\begin{table*}[t!]
\centering
\caption{\textmd{Link prediction results on the FB15k and WN18 datasets for mean reciprocal rank (MRR), and Hits @ n metrics. We evaluate BR-GCN and R-GCN as standalone models and as autoencoder models using HG embedding models as decoders: DistMult (D)~\cite{Distmult}, TransE (T)~\cite{TransE}, HolE (H)~\cite{HolE}, and ComplEx (C)~\cite{ComplEx}. BR-GCN$_{embedding}$ is an ensemble model with a trained BR-GCN model and a separately trained embedding model. Similarly for R-GCN$_{embedding}$. Experiments are run using the Deep Graph Library (DGL)~\cite{dgl}, and the PyTorch Geometric framework~\cite{pytorch-geometric}, on an NVIDIA Tesla V100 GPU cluster. Averages are reported for 10 runs.}}
\begin{tabular}{@{}ccc|ccc||cc| ccc@{}}
\hline
      & \multicolumn{5}{c}{FB15k}                             & \multicolumn{5}{c}{WN18}                            \\ \hline \cmidrule(lr){2-6} \cmidrule(l){7-11}
      & \multicolumn{2}{c}{\texttt{MRR}} & \multicolumn{3}{c}{\texttt{Hits @}} & \multicolumn{2}{c}{\texttt{MRR}} & \multicolumn{3}{c}{\texttt{Hits @}} \\ \cmidrule(lr){2-3} \cmidrule(lr){4-6} \cmidrule(lr){7-8} \cmidrule(l){9-11}
Model & Raw      & Filtered     & 1       & 3      & 10      & Raw      & Filtered     & 1       & 3      & 10      \\ \hline
R-GCN  			& 0.251 & 0.651 & 0.541 & 0.736 & 0.825 & 0.553 & 0.814 & 0.686 & 0.928 & 0.955 \\
BR-GCN  			& 0.255 & 0.662 & 0.564 & 0.748 & 0.829 & 0.557 & 0.814 & 0.691 & 0.928 & 0.956 \\
R-GCN$_{D}$			& 0.262 & 0.696 & 0.601 & 0.760 & 0.842 & 0.561 & 0.819 & 0.697 & 0.929 & 0.964  \\ 
R-GCN$_{T}$  			& 0.252 & 0.651 & 0.543 & 0.738 & 0.828 & 0.554 & 0.815 & 0.681 & 0.928 & 0.956 \\
R-GCN$_{H}$  			& 0.257 & 0.659 & 0.556 & 0.744 & 0.839 & \textbf{0.567} & 0.822 & 0.699 & 0.933 & 0.966 \\
R-GCN$_{C}$  			& 0.260 & 0.712 & 0.629 & 0.771 & 0.845 & 0.565 & 0.822 & 0.701 & 0.933 & 0.965 \\
BR-GCN$_{D}$  			& \textbf{0.265} & 0.703 & 0.646 & 0.782 & \textbf{0.851} & 0.564 & 0.825 & 0.700 & \textbf{0.934} & 0.966 \\
BR-GCN$_{T}$  			& 0.254 & 0.655 & 0.544 & 0.740 & 0.829 & 0.559 & 0.815 & 0.684 & 0.928 & 0.960 \\
BR-GCN$_{H}$  			& 0.258 & 0.661 & 0.560 & 0.748 & 0.840 & \textbf{0.567} & 0.823 & \textbf{0.702} & \textbf{0.934} & \textbf{0.969} \\
BR-GCN$_{C}$  			& 0.264 & \textbf{0.725} & \textbf{0.668} & \textbf{0.785} & \textbf{0.851} & 0.566 & \textbf{0.829} & \textbf{0.702} & \textbf{0.934} & 0.966 \\
\hline
\end{tabular}
\label{table:link-results}
\end{table*}
\vspace{0mm}

\paragraph{\textbf{Datasets}}
To ensure fair comparison against reported results of baseline models, we evaluate BR-GCN on the commonly used node classification heterogeneous datasets in the Resource Description Framework (RDF) format~\cite{benchmark-datasets, RDF}: AIFB, MUTAG, BGS, and AM. Consistent with~\cite{rgcn}, relations used to create entity labels: \textsl{employs} and \textsl{affiliation} (AIFB), \textsl{isMutagenic} (MUTAG), \textsl{hasLithogenesis} (BGS), and \textsl{objectCategory} and \textsl{material} (AM) are removed. The reader is referred to the Appendix for dataset details. 

\paragraph{\textbf{Results}}
The experimental results of node classification are reported in Table~\ref{table:entity-classificaton-results}, which use the benchmark splits from~\cite{rgcn}. We evaluate against state-of-the-art GNNs with two hidden layers for fairness of comparison: HAN~\cite{HAN}, R-GCN~\cite{rgcn}, GAT~\cite{GAT}, and GIN~\cite{GIN}, in addition to BR-GCN variant models: BR-GCN-node being a uni-level attention model using BR-GCN's node-level attention, BR-GCN-relation being a uni-level attention model using BR-GCN's relation-level attention, and BR-GCN is our model, described in Section~\ref{BR-GCN-architecture}. We use the graph defined by all meta-paths for evaluating HAN, since the authors do not indicate how the pre-defined meta-paths are selected. BR-GCN outperforms prior state-of-the-art models on the benchmark datasets and in comparison to BR-GCN-node and BR-GCN-relation. Furthermore, HAN, another bi-level attention model, achieves test accuracy that is generally higher on all datasets compared to the uni-level attention models of GAT, BR-GCN-node, and BR-GCN-relation. This suggests that bi-level attention may be more effective than uni-level attention. While there is insufficient evidence to determine whether relation-level or node-level attention is more important for learning embeddings, bi-level attention seems to leverage the information of both uni-level attentions using an effective weighted aggregation. Hyperparameter values for BR-GCN models are reported in the Appendix.  

\subsection{Link Prediction}
Denote $\mathcal{E^{\prime}}$ to be the incomplete subset of edges $\mathcal{E}$ in the HG. Link prediction involves assigning confidence scores $\alpha$ to \textsl{(h,r,t)} to determine how likely those predicted edges belong to $\mathcal{E}$, or the true relations. We construct graph auto-encoder models, with BR-GCN used as the encoder, and HG embedding models used as the decoder, for this task. We use the approach of~\cite{rgcn} to train and evaluate our model. As such, our models use negative sampling, with $\omega$ being the number of negative samples per observed example. Negative sampling is constructed by randomly corrupting entities of the observed example. We use cross-entropy for the loss function, where observable triples are scored higher than negative triples, with parameters learned using the Adam Optimizer. The loss function is calculated as: 
\begin{equation}
    L = c \times \sum_{(h,r,t,y) \in T}y \mathrm{log}l(\alpha) + (1 - y)\mathrm{log}(1 - l(\alpha))  
    \label{eq:link-pred-loss}
\end{equation}
\begin{equation}
    c = -\frac{1}{(1 + \omega)|\mathcal{E^{\prime}}|}
\end{equation}
\begin{equation}
    y=
\begin{cases}
1 & \text{if positive triples}\\
0 & \text{if negative triples}
\end{cases}
\label{eq:y-indicator}
\end{equation}
with $T$ being the total set of real and corrupted triples, $\alpha$ being the confidence score, $\mathcal{E^{\prime}}$ being the incomplete subset of edges in the HG, $\omega$ being the number of negative samples per observed example, $y$ being an indicator variable defined in Eq.~\ref{eq:y-indicator}, and $l(\cdot)$ being the logistic sigmoid activation.

\paragraph{\textbf{Datasets}}
To ensure fair comparison against reported results of baseline models, we evaluate BR-GCN on the commonly used link prediction heterogeneous datasets of FB15k~\cite{FB15k}, WN18~\cite{WIN18RR}, and FB15k-237~\cite{FB15k-237}, a reduced version of FB15k. Results are based on splits from~\cite{TransE}. The reader is referred to the Appendix for properties of the datasets. 

\vspace{-1mm}
\paragraph{\textbf{Results}}
We use mean reciprocal rank (MRR) and Hits @ n as evaluation metrics, computed in a raw and filtered setting. The reader is referred to~\cite{TransE} for details. The same number of negative samples, with $w = 1$, are utilized to make the datasets comparable. We evaluate BR-GCN and R-GCN as standalone models and as autoencoder models as in~\cite{rgcn}, using the following HG embedding models as decoders: DistMult (D)~\cite{Distmult}, TransE (T)~\cite{TransE}, HolE (H)~\cite{HolE}, and ComplEx (C)~\cite{ComplEx}. BR-GCN$_{embedding}$ is an ensemble model with a trained BR-GCN model and a separately trained embedding model:
$\alpha_{\mathrm{BR-GCN}_{embedding}} = \beta \times \alpha_{\mathrm{BR-GCN}} + (1 - \beta) \times \alpha_{embedding}$, $\beta = 0.4$. Similarly, for R-GCN$_{embedding}$. See Table~\ref{table:link-results} experiment results on FB15k and WN18. The best BR-GCN models outperform R-GCN models. In general, the ComplEx model achieves promising results perhaps since it explicitly models asymmetry in relations. Experiment results for FB15k-237 and further performance analysis are in the Appendix.  

\subsection{Ablation Studies}
We conduct experiments to determine the quality of relation-level attention and graph-structure learned by BR-GCN. We modify the AM dataset to contain the following relations each with cummulative 10\% splits: (1) relations considered by random selection, (2) relations with the highest relation-level attention weights learned by BR-GCN, and (3) relations with the lowest relation-level attention weights learned by BR-GCN. The graph structure resulting from relations of (2) yields the highest test accuracy on all splits of the AM dataset compared to relations from (1) or (3). Furthermore, (3) yields the lowest test accuracy on all splits of the AM dataset. This suggests that BR-GCN's relation-level attention may be useful for selectively identifying important graph components. The Appendix further details these experiments and suggests how BR-GCN's graph structure learning may be transferable to enhance state-of-the-art baseline GNN models. 

\begin{figure}[hbt!]
    \centering
    \includegraphics[width=9.5cm, height=6cm]{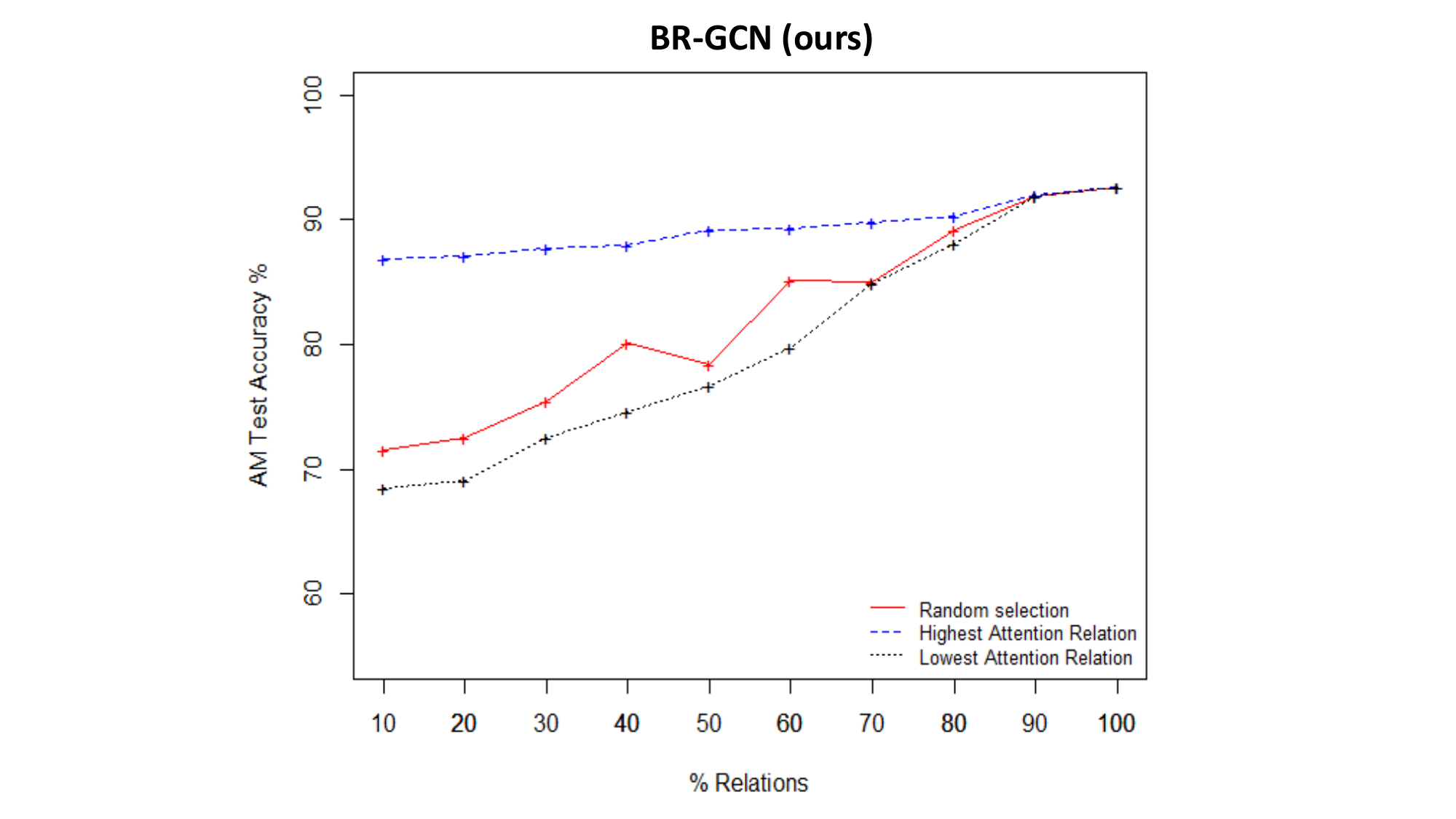}
    \caption{\textmd{Relation-level attention and graph structure learned by BR-GCN.}}
    \label{fig:BR-GCN-ablation}
\end{figure}

\section{Theoretical Analysis}
\begin{table}[htp]
  \caption{\textmd{Runtime complexity for BR-GCN and primary baseline models for computing $\mathbf{h}_{i}^{(l+1)}$.}}
  \label{table:runtime-complexity}
  \centering
  \begin{tabular}{cc}
  \hline
 Model & Runtime Complexity\\ [0.5 ex]
\hline
 HAN & $\mathrm{O}(P \cdot d^{(l)}) + \mathrm{O}(|V_{P}| \cdot d^{3(l)})$ \\[0.5ex] 
 R-GCN & $O(|R_{i}| \cdot |N_{i}^{r}|^{2} \cdot d^{(l+1)} \cdot d^{(l)})$ \\[0.5ex] 
 GAT & $O(K \cdot |N_{i}^{r}| \cdot d^{2(l)})$ \\[0.5ex] 
 BR-GCN-node & $O(|R_{i}| \cdot |N_{i}^{r}| \cdot d^{(l+1)} \cdot d^{(l)})$ \\[0.5ex] 
 BR-GCN-relation & $O(|R_{i}| \cdot |N_{i}^{r}| \cdot d^{(l+1)} \cdot d^{(l)})$ \\[0.5ex] 
 BR-GCN (ours) & $O(d^{2(l)}) + O(|R_{i}| \cdot |N_{i}^{r}| \cdot d^{(l)})$ \\ [0.5ex] 
 \hline
 \end{tabular}
\end{table}
\begin{table}[htp]
  \caption{\textmd{Memory complexity for BR-GCN and primary baseline models for computing $\mathbf{h}_{i}^{(l+1)}$.}}
  \label{table:memory-complexity}
  \centering
  \begin{tabular}{ccc}
  \hline
 Model & Memory Complexity\\ [0.5 ex]
\hline
 HAN  & $O(d^{(l)}\times (K+|R|))$\\[0.5ex] 
 R-GCN & $O(|R_{i}| \cdot d^{(l+1)} \cdot d^{(l)})$ \\[0.5ex] 
 GAT & $O(d^{(l)} \times (K + |R_{i}|))$ \\[0.5ex] 
 BR-GCN-node & $O(|R_{i}| \cdot d^{(l+1)} \cdot d^{(l)})$ \\[0.5ex]
 BR-GCN-relation & $O(d^{(l+1)})$ \\[0.5ex] 
 BR-GCN (ours) & $O(|R_{i}| \cdot d^{(l)})$ \\ [0.5ex] 
 \hline
 \end{tabular}
\end{table}

Table~\ref{table:runtime-complexity} shows the runtime complexity for BR-GCN and the primary baseline models, and Table~\ref{table:memory-complexity} shows the memory complexity for BR-GCN and the primary baseline models. $d^{(l)}$ and $d^{(l+1)}$ are the node dimensions at the $l$-th and $(l+1)$-th layers respectively, $R_{i}$ is the set of relations in the neighborhood context, $R$ is the set of relations in the HG, $N_{i}^{r}$ is the set of nodes in the neighborhood context, $K$ is the number of attention heads, $P$ is the set of pre-defined meta-paths used in HAN, and $V_{p}$ is the set of nodes contained in the pre-defined meta-paths. The reader is referred to~\cite{rgcn, GAT, HAN} for descriptions of the baseline GNN models. BR-GCN is comparable in runtime and memory complexity to the state-of-the-art neural architectures of HAN, R-GCN, and GAT, suggesting that it is a scalable architecture. 
\section{Conclusions and Future Work}
\label{conclusions}

We present a generalized framework for computing bi-level attention and discuss our novel bi-level attention model, BR-GCN. Our best BR-GCN model effectively leverages attention and graph sparsity as suggested by experiment results against state-of-the-art baseline models of HAN, R-GCN, GAT, GIN, BR-GCN-node, and BR-GCN-relation. On node classification, BR-GCN outperforms baselines from 0.29\% to 14.95\%, and on link prediction, BR-GCN outperforms baselines from 0.02\% to 7.40\%. As GNNs and attention-based neural architectures have been widely applied to numerous domains, by advancing these architectures, BR-GCN also has the potential to further advance knowledge discovery in these domains. Domain applications of BR-GCN include social networks, medical informatics, software development, and natural language processing. For future work, we plan to investigate applying BR-GCN to enhance the inference tasks of question-answering for text and software, graph similarity detection, as well as to benefit domain-specific tasks such as code recommendation and bug detection in software development, and disease prediction and prognosis in medical informatics.

\balance
\newpage
\paragraph{\textbf{Acknowledgements}}

We would like to thank Yunsheng Bai for helpful discussions, comments, and corrections. This project is supported by the National Science Foundation (NSF) Research Traineeship program, through the NSF \textbf{M}od\textbf{E}ling and u\textbf{N}ders\textbf{T}anding human behavi\textbf{OR} (MENTOR) Fellowship (Grant No. \#1829071).

\bibliographystyle{ACM-Reference-Format}
\def\bibfont{\large}
\bibliography{main}

\newpage
\appendix 

\section*{APPENDIX}

\section{Source Code and Environment}
The source code for our work can be found at:
\url{https://github.com/roshnigiyer/BR-GCN} \vspace{1mm}

Details for running the source code are present in the $\mathrm{README.md}$ file, and environment details are in the $\mathrm{requirements.txt}$ file.

\begin{figure*}[h]
    \centering
    \includegraphics[width=14cm]{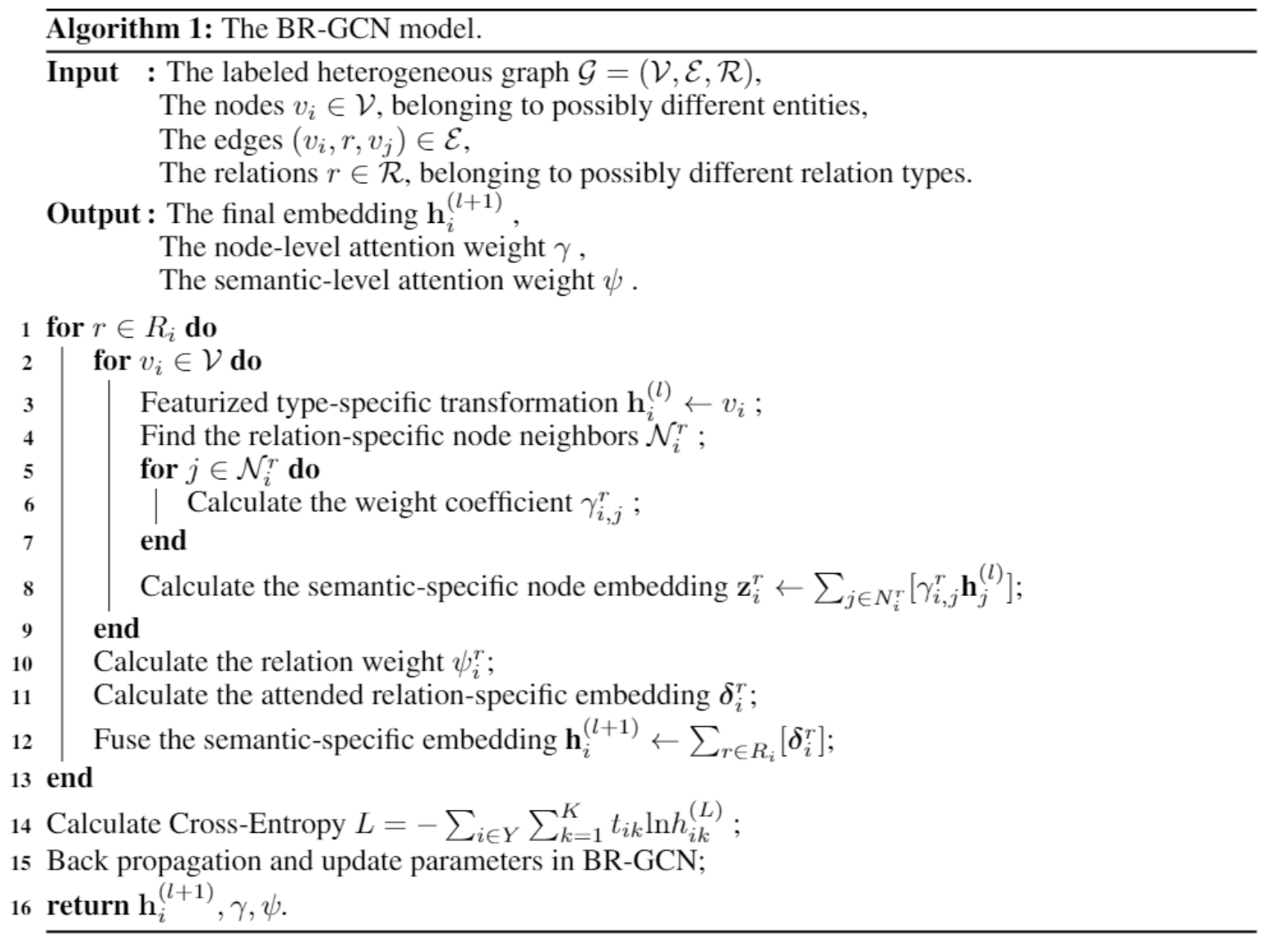}
    \label{algorithm:BR-GCN}
\end{figure*}
\begin{figure*}[h]
    \centering
    \includegraphics[width=15cm]{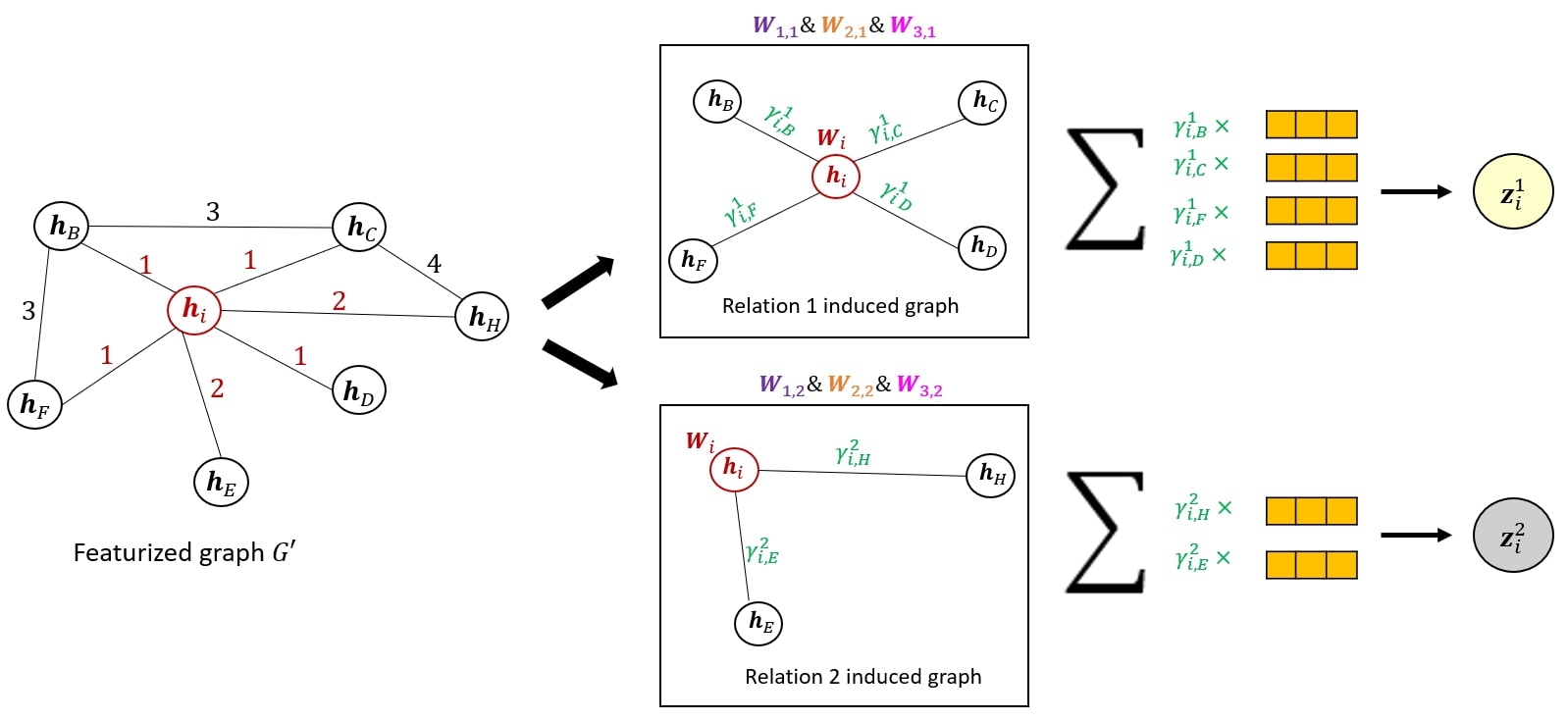}
    \includegraphics[width=9cm]{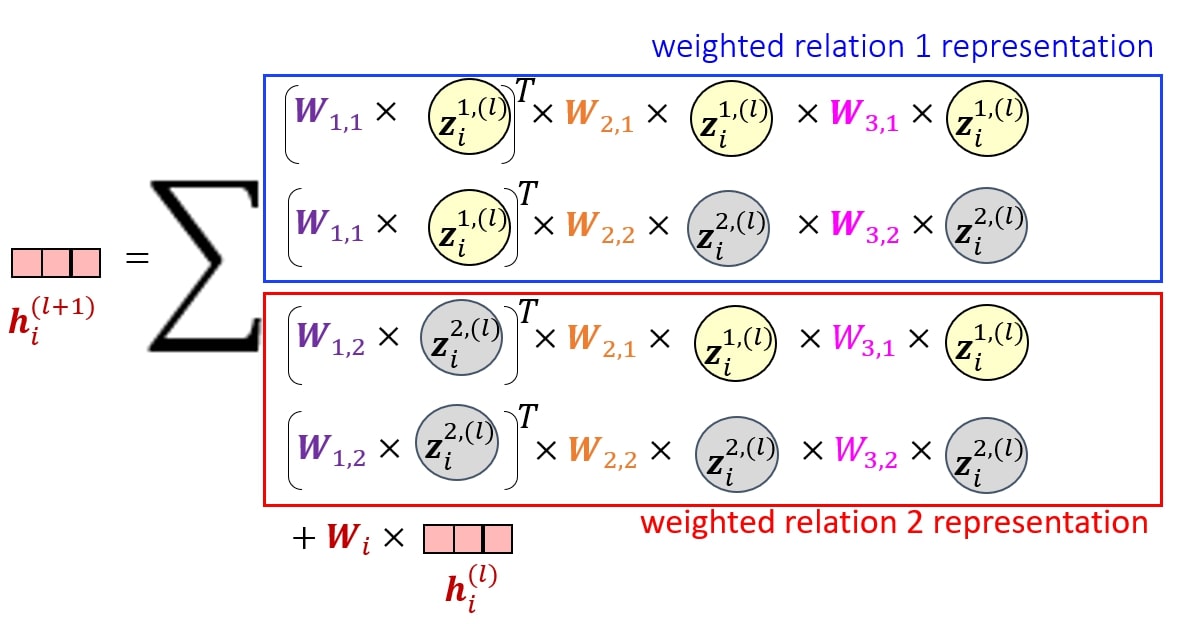}
    \caption{\textmd{Bi-level attention visualization walkthrough example. The featurized graph $G^{\prime}$ of node $i$ denotes relations in the neighborhood context for node $i$ in red. An induced graph for each relation in the neighborhood context is created and node-level attention for each induced graph is learned to form relation-specific embeddings $\textbf{z}_{i}^{1}$, and $\textbf{z}_{i}^{2}$ for relations 1 and 2. The relation-level attention mechanism to learn the final node embedding $\textbf{h}_{i}^{(l+1)}$ and the corresponding weighted relation representations of relations 1 and 2 are shown.}}
    \label{fig:BR-GCN-inter-intra2-walkthrough-ex}
\end{figure*}

\vspace{2mm}
The overall process of BR-GCN is described in Algorithm 1.
A walkthrough example of BR-GCN's architecture is detailed in Figure~\ref{fig:BR-GCN-inter-intra2-walkthrough-ex}.

\section{Dataset Description}
We evaluate node classification on four datasets: AIFB, MUTAG, BGS, and AM. We evaluate link prediction on three datasets: FB15k, WN18, and FB15k-237. We evaluate ablation studies on the AM dataset. The description of these datasets are found below. 
\label{appendix-datasets}

\paragraph{\textbf{AIFB}}
The AIFB dataset is a social network dataset pertaining to the Institute for Applied Informatics and Formal Description Methods at the Karlsruhe Institute of Technology. It includes the relationships between persons (e.g., Professors, Students), research topics, projects, publications etc. AIFB has 8,285 entities, 45 relations, and 29,043 edges. 176 of the entities have labels and are to be classified into 4 classes. 

\paragraph{\textbf{MUTAG}}
The MUTAG dataset is a biological dataset that contains information about molecules that are potentially carcinogenic. MUTAG has 23,644 entities, 23 relations, and 74,227 edges. 340 of the entities have labels and are to be classified into 2 classes. 

\paragraph{\textbf{BGS}}
The BGS dataset is a geological dataset that contains information about rock units. BGS has 333,845 entities, 103 relations, and 916,199 edges. 146 of the entities have labels and are to be classified into 2 classes. 

\paragraph{\textbf{AM}}
The AM dataset contains information about artifacts from the Amsterdam Museum. AM has 1,666,764 entities, 133 relations, and 5,988,321 edges. 1,000 of the entities have labels and are to be classified into 11 classes. 

\paragraph{\textbf{FB15k}}
The FB15k dataset is a subset of Freebase, a highly multi-relational collaborative HG. FB15k has 14,951 entities, and 1,345 relations. 483,142 edges are used for training, 50,000 edges are used for validation, and 59,071 edges are used for testing.

\paragraph{\textbf{WN18}}
The WN18 dataset is a subset of WordNet, a multi-relational lexical HG for the English language. WN18 has 40,943 entities, and 18 relations. 141,442 edges are used for training, 5,000 edges are used for validation, and 5,000 edges are used for testing.

\paragraph{\textbf{FB15k-237}}
The FB15k-237 dataset is a reduced dataset of FB15k with inverse triplet pairs removed. Triplet and inverse triplet pairs are denoted as: $t = (e_{1}, r, e_{2})$ and $t^{-1} = (e_{2}, r^{-1}, e_{1})$. FB15k-237 has 14,541 entities, and 237 relations. 272,115 edges are used for training, 17,535 edges are used for validation, and 20,466 edges are used for testing.

\section{Experiments}
In this section we describe further experimental details for BR-GCN and baseline models for node classification, link prediction, and ablation studies. 

\subsection{Model Hyperparameters}
\label{subsection:node-classification}
We proceed to summarize the hyperparameter values used for BR-GCN models for node classification experiments. The hyperparameters used for BR-GCN models for link prediction experiments and ablation studies are the same as the hyperparameter values for
\vspace{2mm}BR-GCN models for the AM dataset. 

The learning rate for BR-GCN-node for the datasets of AIFB, MUTAG, BGS, and AM are \{0.010, 0.001, 0.001, 0.001\} respectively, with the $l2$ penalty as 0 and the \# hidden units as 16 for all datasets. The \# basis functions are \{6, 1, 0, 2\} respectively, and the \# epochs are \{70, 90, 70, 80\} respectively. The dropout rate is \{0.6, 0.4, 0.0, 0.6\} respectively. The negative slope for $\mathrm{LeakyReLU}(\cdot)$ activation is \{0.6, 0.8, 0.4, 0.8\} \vspace{2mm} respectively. 

The learning rate for BR-GCN-relation for the datasets of AIFB, MUTAG, BGS, and AM are \{0.010, 0.010, 0.050, 0.001\} respectively, with the $l2$ penalty as \{0, 0, 0, $5 \times 10^{-4}$\} respectively. The \# hidden units are 16 for all datasets. The \# basis functions are \{2, 0, 4, 2\} respectively, and the \# epochs are \{70, 75, 85, 85\} \vspace{2mm} respectively.

The learning rate for BR-GCN for the datasets of AIFB, MUTAG, BGS, and AM are \{0.050, 0.010, 0.005, 0.010\} respectively, with the $l2$ penalty being \{0, $5 \times 10^{-4}$, 0, 0\} respectively. The \# hidden units are 16 for all datasets. The \# basis functions are \{0, 0, 1, 0\} respectively, and the \# epochs are \{85, 90, 95, 100\} respectively. The dropout rate is \{0.4, 0.2, 0.6, 0.6\} respectively. The negative slope for $\mathrm{LeakyReLU}(\cdot)$ activation is \{0.2, 0, 0.4, 0\} respectively.

\subsection{Link Prediction}

\begin{table*}[htp!]
\caption{\textmd{Link prediction results on the FB15k-237 dataset, a reduced version of FB15k with problematic inverse relation pairs removed for mean reciprocal rank (MRR), and Hits @ n metrics. We evaluate BR-GCN auto-encoder models against R-GCN auto-encoder models~\cite{rgcn} using the following HG embedding models as decoders: DistMult (D)~\cite{Distmult}, TransE (T)~\cite{TransE}, HolE (H)~\cite{HolE}, and ComplEx (C)~\cite{ComplEx}. R-GCN$_{embedding}$ denotes R-GCN as the encoder combined with the specific embedding model as the decoder. Similarly, for BR-GCN$_{embedding}$. Experiments are run using the Deep Graph Library (DGL)~\cite{dgl}, and the PyTorch Geometric framework~\cite{pytorch-geometric}.}}
\centering
\begin{tabular}{ccc|ccc}
\hline
                                  & \multicolumn{2}{c}{\texttt{MRR}} & \multicolumn{3}{c}{\texttt{Hits @}} \\ \cmidrule{2-3} \cmidrule{4-6}
Model                             & Raw      & Filtered     & 1       & 3       & 10     \\ \hline
R-GCN               & 0.158       & 0.248 & 0.153     & 0.258     & 0.414     \\
BR-GCN  			& 0.160 & 0.249 & 0.160 & 0.261 & 0.418 \\
R-GCN$_{D}$                     & 0.156 & 0.249 & 0.151     & 0.264     & 0.417 \\
R-GCN$_{T}$  			& 0.161 & 0.258 & 0.159 & 0.274 & 0.421 \\
R-GCN$_{H}$  			& 0.159 & 0.257 & 0.156 & 0.272 & 0.420 \\
R-GCN$_{C}$  			& 0.158 & 0.255 & 0.152 & 0.268 & 0.419 \\
BR-GCN$_{D}$  			& 0.157 & 0.251 & 0.255 & 0.265 & 0.419 \\
BR-GCN$_{T}$  			& \textbf{0.163} & \textbf{0.261} & \textbf{0.164} & \textbf{0.275} & \textbf{0.423} \\
BR-GCN$_{H}$  			& 0.161 & 0.258 & 0.158 & 0.272 & 0.422 \\
BR-GCN$_{C}$  			& 0.160 & 0.259 & 0.153 & 0.268 & 0.420 \\
\hline
\end{tabular}
\label{table:fb15k-237}
\end{table*}

To ensure fair comparison against reported results of baseline models, we evaluate BR-GCN on the commonly used link prediction heterogeneous datasets of FB15k, WN18, and FB15k-237. See Section~\ref{appendix-datasets} for dataset descriptions. Table 3 presents our experiment results for FB15k and WN18. Table~\ref{table:fb15k-237} presents our experiment results for \vspace{2mm} FB15k-237. 

We find that the best BR-GCN models outperform R-GCN models on all three datasets. The FB15k and WN18 datasets are well-connected and therefore nodes can learn important information from their neighborhood contexts. As such, the ComplEx decoder model seems to achieve promising results on these datasets perhaps since it explicitly models asymmetry in relations. This may enable the model to learn more nuances in information that other HG embedding models fail to capture. FB15k-237, however, does not contain as much important local information. Therefore, HG embedding models are not 
as useful when being coupled with R-GCN and BR-GCN in FB15k-237 compared to in the FB15k and WN18 datasets. The TransE embedding model seems to be most effective in FB15k-237 perhaps because the embedding relationship between the head and tail entities are linear. Refer to Section~\ref{subsection:node-classification} for details on model hyperparameters.

\subsection{Ablation Studies}

\begin{figure*}[t!]
    \centering
    \includegraphics[width=14cm]{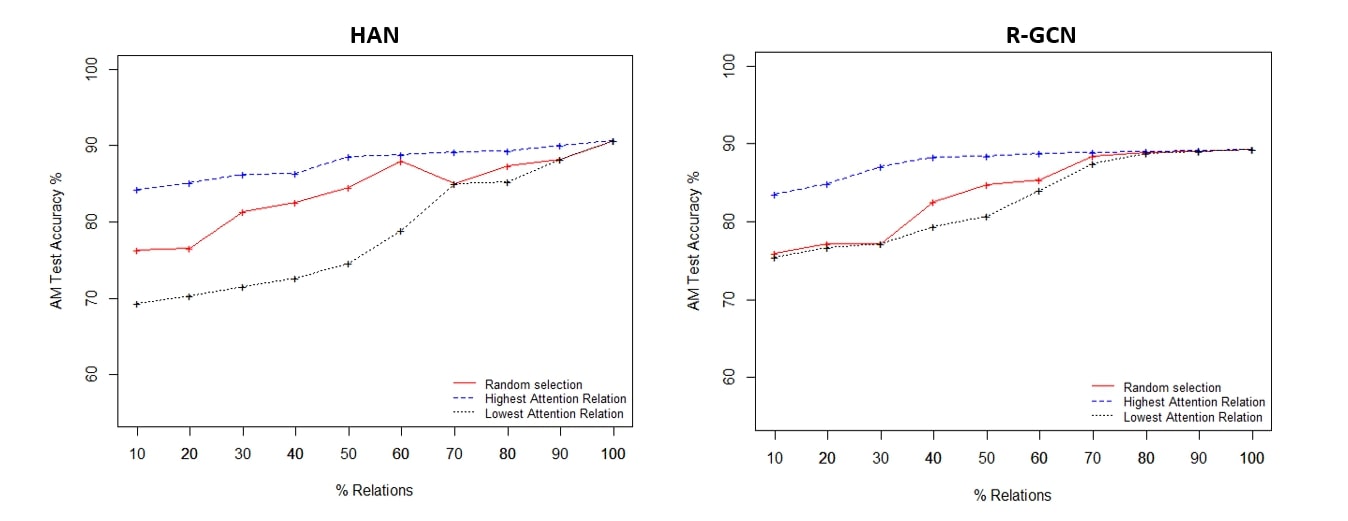}
    \includegraphics[width=14cm]{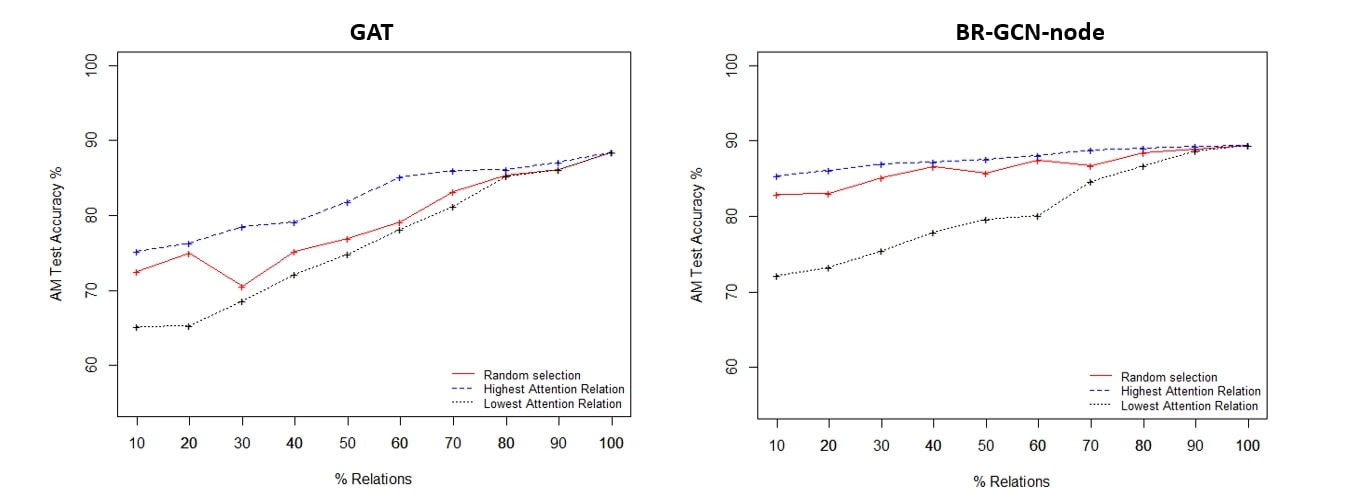}
    \includegraphics[width=14cm]{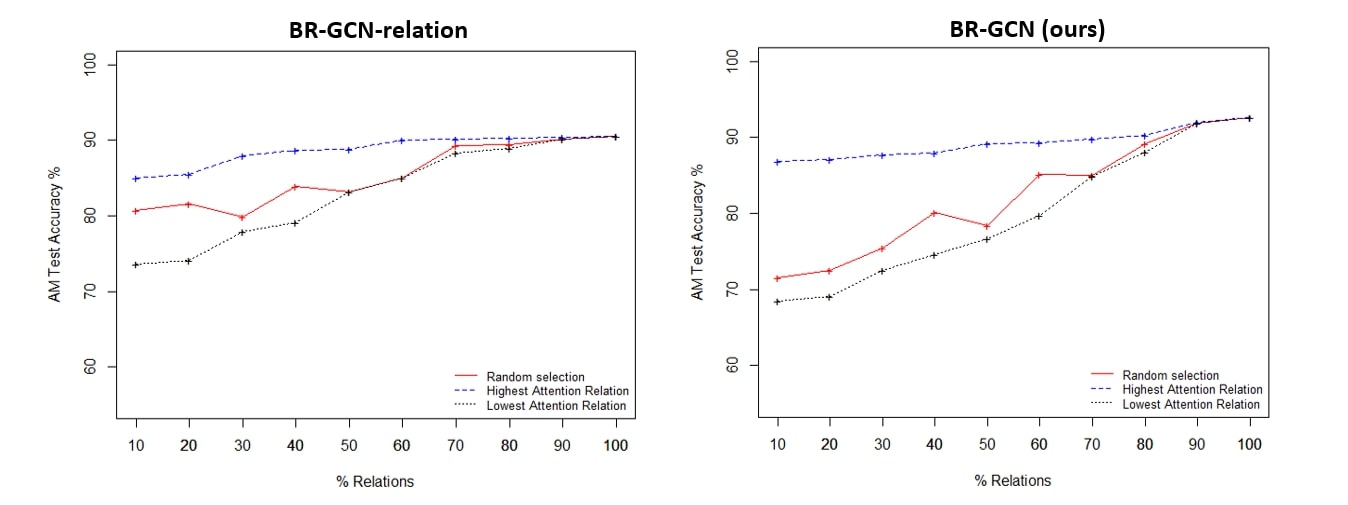}
    \caption{\textmd{Ablation study experiments on primary baseline models of HAN, R-GCN, GAT, and BR-GCN models to evaluate learned relation-level attention and graph structure of BR-GCN.}}
    \label{fig:ablation-study}
\end{figure*}
\begin{table*}[htp]
  \caption{\textmd{Test Accuracy (\%) achieved from the
  top $x$\% attention relations for the AM dataset. The attention of relations is a function of the learned relation-level attention matrix $\psi^{r}$ computed by BR-GCN, defined as $average(\psi^{r})$.}}
  \label{table:semantic-attention-experiment}
  \centering
  \begin{tabular}{cccc}
  \hline
 Model & Top 10\% & Top 50\% & Top 100\%\\ [0.5 ex]
\hline
 HAN & 84.32 & 88.55 & 90.68\\[0.5ex] 
 R-GCN & 83.61 & 88.47 & 89.29\\[0.5ex] 
 GAT & 75.25 & 81.90 & 88.52\\[0.5ex] 
 BR-GCN-node & 85.38 & 87.63 & 89.45\\[0.5ex] 
 BR-GCN-relation & 84.97 & 88.89 & 90.52\\[0.5ex] 
 BR-GCN (ours) & \textbf{85.91} & \textbf{89.19} & \textbf{92.55}\\ [0.5ex] 
 \hline
 \end{tabular}
\end{table*}

We conduct experiments to determine the quality of relation-level attention and graph-structure learned from BR-GCN. We modify the AM dataset to contain the following types of relations, each with cummulative 10\% splits: (1) relations considered by random selection, (2) relations with the highest corresponding relation-level attention weights learned by BR-GCN, and (3) relations with the lowest corresponding relation-level attention weights learned by BR-GCN. The attention of relations for BR-GCN that we utilize is a function of the learned relation-level attention matrix $\psi^{r}$, which is computed by averaging the matrix elements: $average(\psi^{r})$. The experiment figures for the primary baseline models of HAN, R-GCN, GAT, and BR-GCN models are detailed in \vspace{2mm} Figure~\ref{fig:ablation-study}. 

As shown in Figure~\ref{fig:ablation-study}, the AM graph produced by the highest attention relation results in the highest test accuracy compared to the other metrics in all models.  The AM graph produced by the lowest attention relation results in the lowest test accuracy compared to the other metrics in all models. The test accuracy for the AM graph produced by relations randomly selected are as expected in between the test accuracies of the other two metrics. Since models that do not learn relation-level attention (R-GCN, GAT, BR-GCN-node) still benefit from the graph structure identified by the highest attention relations, this shows that the learning of graph structure identified by relation-level attention may be transferable. Furthermore, as larger \% of relations are considered, all models benefit from learning better graph structure and some models may also learn better attention, resulting in even higher test accuracies.    

Table~\ref{table:semantic-attention-experiment} summarizes the model test accuracies achieved from the AM graph produced by the highest attention relation metric for the top 10\%, 50\%, and 100\% of relations. BR-GCN produces the highest test accuracy in all categories, showing that it may learn the best from graph structure and relation-level attention compared to the primary baseline models.

\end{document}